\documentclass[journal]{IEEEtran}
\usepackage{graphicx}
\usepackage[switch]{lineno}
\usepackage{comment}

\ifCLASSINFOpdf
\else
\fi

\begin{document}
\title{GridTracer: Automatic Mapping of Power Grids using Deep Learning and Overhead Imagery}
%
%
%

\author{Bohao~Huang,~\IEEEmembership{Member,~IEEE,}
        Jichen~Yang,~\IEEEmembership{Member,~IEEE,}
        Artem~Streltsov,~\IEEEmembership{Member,~IEEE,}
        Kyle~Bradbury,~\IEEEmembership{Member,~IEEE,}
        Leslie~M.~Collins,~\IEEEmembership{Senior Member,~IEEE,}
        and~Jordan~Malof,~\IEEEmembership{Member,~IEEE}
}

%

\maketitle

\begin{abstract}
Energy system information valuable for electricity access planning such as the locations and connectivity of electricity transmission and distribution towers – termed the power grid – is often incomplete, outdated, or altogether unavailable. Furthermore, conventional means for collecting this information is costly and limited. We propose to automatically map the grid in overhead remotely sensed imagery using deep learning. Towards this goal, we develop and publicly-release a large dataset ($263km^2$) of overhead imagery with ground truth for the power grid – to our knowledge this is the first dataset of its kind in the public domain. Additionally, we propose scoring metrics and baseline algorithms for two grid mapping tasks: (1) tower recognition and (2) power line interconnection (i.e., estimating a graph representation of the grid).  We hope the availability of the training data, scoring metrics, and baselines will facilitate rapid progress on this important problem to help decision-makers address the energy needs of societies around the world. 
\end{abstract}


%
\IEEEpeerreviewmaketitle

\section{Introduction}
\label{sec:introduction}
Providing access to sustainable, reliable, and affordable energy access is vital to the prosperity and sustainability of modern societies, and it is the United Nation’s Sustainable Development Goal \#7 (SDG7) \cite{Nations2018}. Increased electricity access is correlated with positive educational, health, gender equality, and economic outcomes \cite{Alstone2015}. Ensuring energy access over the coming decades, and achieving SDG7, will require careful planning from non-profits, governments, and utilities to determine electrification pathways to meet rapidly growing energy demand. 

\begin{figure}
    \centering
    \includegraphics[width=0.45\textwidth]{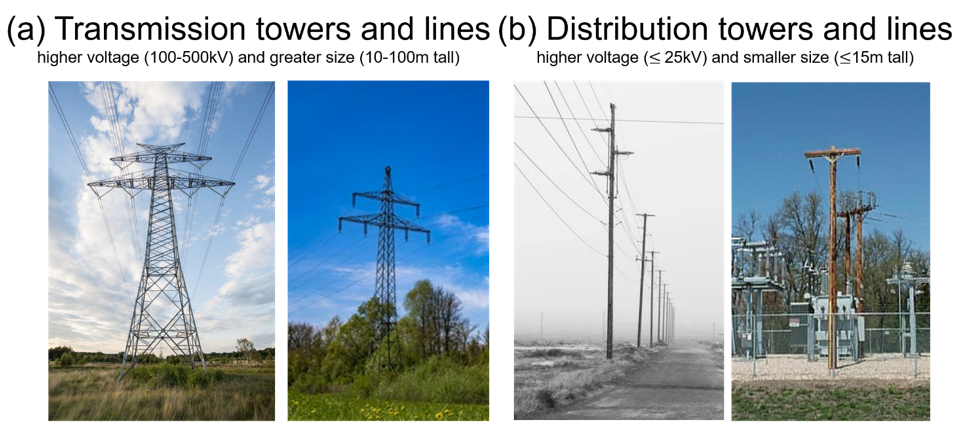}
    \caption{(a) Transmission towers and lines. (b) Distribution towers and lines. Transmission towers carry higher-voltage electricity than distribution towers, and are physically larger.}
    \label{fig:ground_view_towers}
\end{figure}

A crucial resource for this decision-making will be high-quality information about existing power transmission and distribution towers, as well as the power lines that connect them (see Fig. \ref{fig:ground_view_towers}); we collectively refer to these infrastructures as the power grid (PG).  Information about the precise locations of PG towers and line connectivity is crucial for decision-makers to determine cost-efficient solutions for extending reliable and sustainable energy access \cite{Szabo2011}.  For example, this information can be used in conjunction with modeling tools like the Open Source Spatial Electrification Tool (OnSSET) \cite{Mentis2015} to determine the optimal pathway to electrification: grid extension, mini/microgrids, or off-grid systems like solar.

Unfortunately, the PG information available to decision-makers is frequently limited. Existing PG data are often incomplete, outdated, of low spatial resolution, or simply unavailable \cite{DevelopmentSeed2018, arderne2020predictive}. Furthermore, conventional methods of collecting PG data, such as field surveys or collating utility company records, are either costly or require non-disclosure agreements. The importance of this problem and the lack of PG data has recently prompted major organizations such as the World Bank \cite{DevelopmentSeed2018} and Facebook \cite{arderne2020predictive} to investigate solutions to it. In this work we propose to address this problem by using deep learning models to automatically map (i.e., detect and connect) the PG towers detectable in high-resolution color overhead imagery (e.g., from satellite imagery and aerial photography) using deep learning models. 

\subsection{Mapping the grid using overhead imagery}
\label{sec:intro_map}
Recently deep learning models - namely deep neural networks (DNNs) - have been shown to be capable of accurately mapping a variety of objects in color overhead imagery, such as buildings \cite{Huang2018b,demir2018deepglobe} roads \cite{demir2018deepglobe,Bastani2018}, and solar arrays \cite{Malof2016a,yu2018deepsolar,Malof2015}. Since PG towers and lines are often visible in overhead imagery, these results suggest that mapping the PG may be also be feasible.  However, PG mapping presents several unique challenges compared to mapping other objects in overhead imagery.  

The most immediate challenge of PG mapping is the structure of the desired output.  The PG is generally represented as a geospatial graph, where each tower represents a graph node with an associated spatial location, and each PG line represents a connection between two nodes \cite{DevelopmentSeed2018,arderne2020predictive}.  This representation is compact, and well-suited for subsequent use by energy researchers and decision-makers.   Therefore, we require that any automatic recognition model produce a geospatial PG graph as output.  

\begin{figure}
    \centering
    \includegraphics[width=0.45\textwidth]{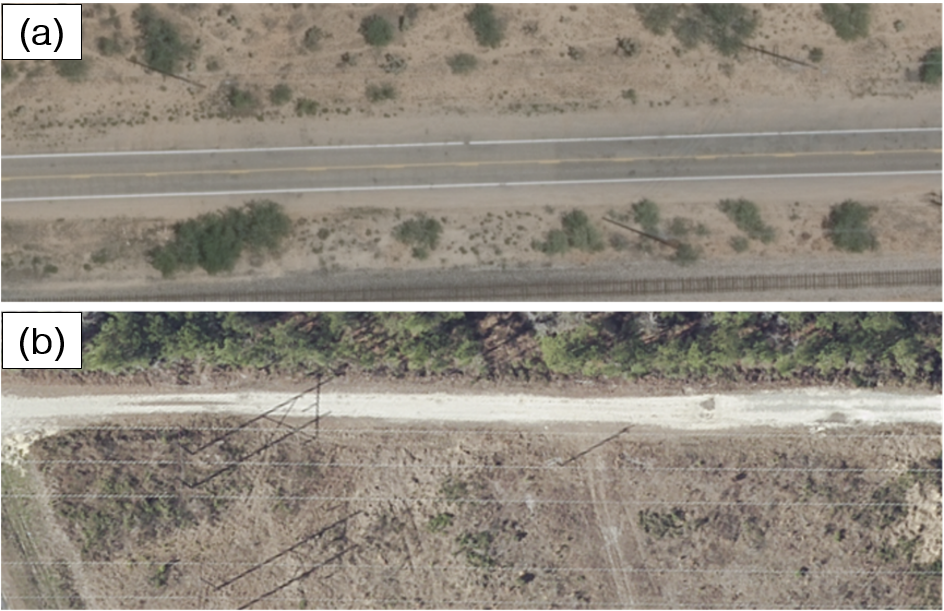}
    \caption{Overhead imagery of power grid towers and lines in two different locations.  (a) Distribution towers and lines in Tucson, Arizona, U.S.A. (b) Transmission towers and lines in Wilmington, North Carolina, U.S.A.}
    \label{fig:towers}
\end{figure}

A second challenge is that PG infrastructure exhibits weak and geographically-distributed visual features in overhead imagery, making the problem both unique and challenging. Fig.  \ref{fig:towers} illustrates the weak visual features of PG infrastructure in representative examples of (a) distribution and (b) transmission PG infrastructure.  Looking closely at Fig. \ref{fig:towers}(a) it is apparent that PG towers exhibit very few visual features (if any), aside from their shadows.  Shadows are useful for detection however their strength and visual appearance varies, and they are not always present (e.g., depending upon time of day).  As  illustrated in Fig. \ref{fig:towers} PG lines typically appear as thin white or black lines that are only intermittently visible due to their varying contrast with the local background (e.g., white lines become faint, or disappear, as they cross pale background).  From Fig. \ref{fig:towers} it is also notable that transmission infrastructure tends to be relatively easy to detect because it is much larger, however, it is also much more rare than distribution infrastructure.

\begin{figure}
    \centering
    \includegraphics[width=0.48\textwidth]{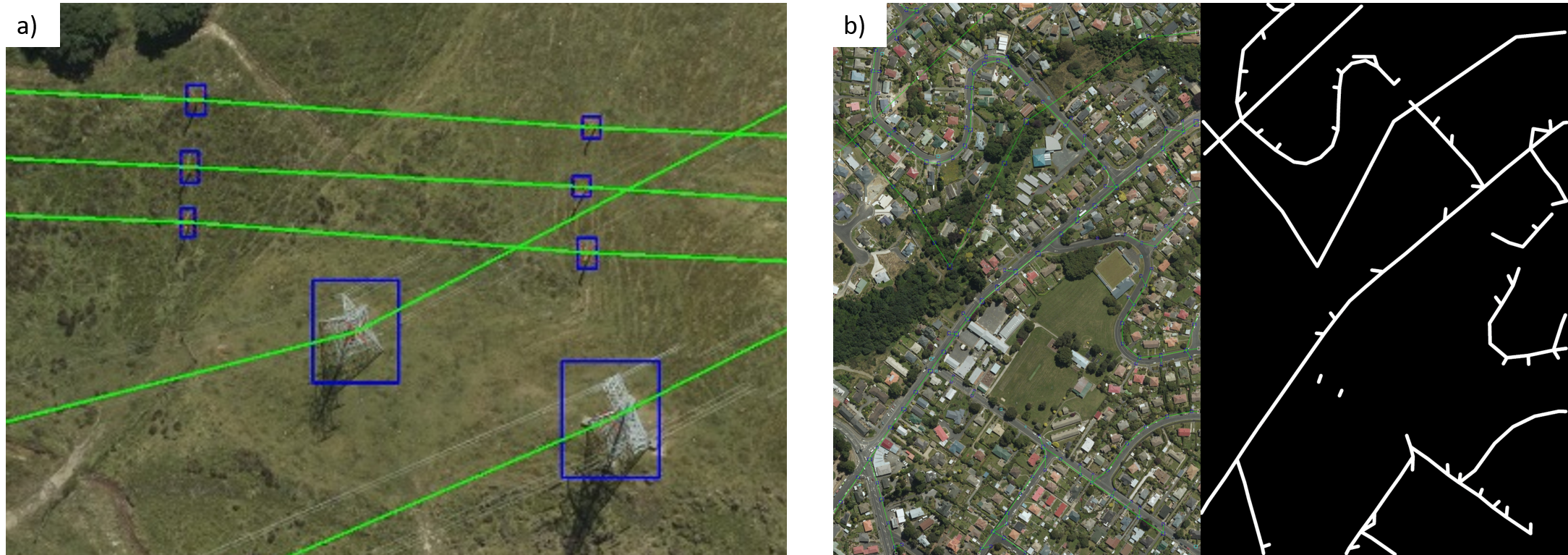}
    \caption{Illustration of the PG inference task, where (a) shows the annotated tower and lines and (b) demonstrates the power grid in larger scale. Note the complexity of the topology in (b).}
    \label{fig:cmp}
\end{figure}

As a result of these two major challenges, PG mapping is a unique and challenging problem for existing visual recognition models. Fig. \ref{fig:cmp}(a) demonstrates the desired output structure of the PG mapping where towers are represented in boxes or nodes and lines are represented in edges. Fig. \ref{fig:cmp}(b) illustrates the large-scale topology of the PG in one region, which might be leveraged for recognition. On the contrast, modern DNNs for example rely primarily upon local visual features for object recognition \cite{li2019scale,luo2016understanding}, making them poorly-suited for PG mapping.  Additionally most existing DNNs do not typically produce output in the form of a geospatial graph.  Some work has recently been conducted for inferring geospatial graphs on overhead imagery for the problem of road mapping \cite{Bastani2018,mattyus2017deeproadmapper}.  Unfortunately, however, these approaches make fundamentally different assumptions about the structure of the underlying graph, and the visual features associated with it, limiting their applicability to the PG mapping problem (see Section \ref{sec:related} for further discussion).   

Therefore, effective PG mapping will require the development of novel models that can address its unique challenges.  This raises a third major challenge of PG mapping: the absence of any publicly-available benchmark dataset to train and validate recognition models.  Furthermore, it is unclear how to score a geospatial graph so that different models can be compared.  Without a publicly-available dataset and scoring metrics, it is not possible to effectively study this problem.  


\subsection{Contribution of this work}
In this work we make two primary contributions and lay the foundations for a practical PG mapping approach with overhead imagery.  First, we develop and publicly-release a dataset of satellite imagery encompassing an area of $263km^2$ across seven cities: two cities in the U.S. and five cities in New Zealand.  In this work we employ imagery with a 0.3m ground sampling distance.  Our primary motive for choosing this resolution is because it is the highest resolution that is also widely employed for research on object recognition in overhead imagery \cite{demir2018deepglobe,Bastani2018,Maggiori2017,Huang2018b}.  Imagery at this resolution is also commercially available across the globe, making any techniques developed using it widely applicable.  Our dataset includes both imagery and corresponding hand annotations of transmission and distribution towers (as rectangles) and the power lines connecting them (as line segments), making it possible to train and validate deep learning approaches for PG mapping.  We also perform additional tests to evaluate the quality of the hand annotations.   To enable future benchmark testing, we describe a data handling procedure for training and validating models, and we propose metrics to score the graphical output of the models.      

Our second contribution is a novel deep model – termed \textit{GridTracer} – that makes a first step towards addressing the unique challenges of PG mapping in overhead imagery. \textit{GridTracer} effectively splits PG mapping into three simpler steps: tower detection, line segmentation, and PG graph inference.  In tower detection we use a deep object detection model to identify the location and size (e.g., with a bounding box) of individual PG towers. In line segmentation we use a deep image segmentation model to generate a pixel-wise score indicating the likelihood that a PG line is present.  In PG graph inference, we integrate the output of steps one and two over large geographic regions, with the goal of estimating which towers are likely to be connected by power lines. The final output of \textit{GridTracer} is a geospatial graph, in which graph nodes represent PG tower locations, and graph edges (node connections) represent PG lines.  

As we discuss in Section \ref{sec:related}, existing DNN-based approaches are not suitable to solve the PG mapping problem, and therefore we propose \textit{GridTracer} as a baseline for future work. To this end, we use our new benchmark dataset to comprehensively study the performance of \textit{GridTracer}, including ablation studies to analyze the major designs and hyperparameter choices of \textit{GridTracer}.  We also compare \textit{GridTracer}'s performance to human-level PG mapping accuracy on our benchmark to provide insights on the level of PG mapping accuracy that may ultimately be achievable, and thereby how much further \textit{GridTracer} could be improved with further research.  We hope the availability of these resources (e.g., data, scoring metrics, and baseline) and analyses will facilitate rapid progress on this important problem to help decision-makers address the energy needs of societies around the world.  



\section{Related Work}
\label{sec:related}
\textbf{Mapping the power grid in remotely-sensed data.} Until recently, the majority of work related to PG analysis using remote sensing techniques focused around identifying vegetation encroachment for monitoring known transmission lines \cite{Matikainen2016, Ahmad2013, Kobayashi2014}. More recently, Facebook \cite{arderne2020predictive} and Development Seed (in partnership with the World Bank) \cite{DevelopmentSeed2018} have both proposed approaches to map PGs in overhead imagery within the past year.

Facebook’s approach uses nighttime lights imagery to identify distribution lines (medium voltage). This approach maps grid connectivity using VIIRS 750m-resolution day/night band nighttime lights imagery to identify and connect communities using electricity. Since nighttime light data is not perfectly correlated with energy access, especially in regions with lower electricity access rates, such an approach could potentially be less accurate in locations where it would be most beneficial for extending electricity access. Additionally, the 750m-resolution of the underlying data prevents individual towers and lines from being directly observed. This approach culminated in the release of a medium-voltage transmission dataset \cite{arderne2020predictive}, which, while global in scope, only made PG estimates at a resolution of one estimate per square kilometer, and reported their performance in a scale no smaller than $1km^2$; both of which are coarser than what is needed for many types of electricity access planning.  

A summary of the approach proposed by Development Seed  is published online, along with software and their output PG mapping dataset\footnote{https://github.com/developmentseed/ml-hv-grid-pub}.  Their approach relies on identifying high voltage transmission infrastructure in color overhead imagery, making it more similar to our proposed approach.  However, because their approach uses a human in the loop, it will be costly to scale it to large geographic areas, or utilize their approach for repeated PG mapping over time. 
   
Our work builds on the foundation laid by each of these approaches, seeking to achieve a high resolution mapping of the PG (i.e., individual towers and connections), and to do so with a fully-automated and scalable pipeline, akin to that of Facebook’s approach.  Furthermore, and in contrast to both existing approaches, we map both transmission and distribution PG infrastructure, providing more comprehensive support for electricity access planning.  Our experimental results also include comprehensive analysis of the proposed \textit{GridTracer} model; providing support for its overall design, its performance under different deployment scenarios, and its sensitivity to hyperparameter settings.

\textbf{Object detection in overhead imagery.} One of the first components of our approach is to identify the PG towers (or poles); for this step, we employ object detection. Object detection algorithms identify and localize the objects of interest within a given image, typically by placing a bounding box around the object. Recent state-of-the-art object detection algorithms employ deep neural networks (DNNs). The studies \cite{lin2017focal, ren2015faster, he2017mask} first use DNNs to propose regions of interest (RoI), after which a second neural network is used to (i) classify and (ii) refine the bounding boxes identified by the first-stage network. This two-stage process generally yields higher accuracy on benchmark problems \cite{lin2014microsoft}. Based upon this approach, \cite{Yang2018a, Li2019a, azimi2018towards} have developed detectors specifically for the overhead imagery to address the specific challenges in remote sensing including the various scales of the objects as well as arbitrary orientation of their bounding boxes \cite{Yang2018a}. However, to our best knowledge there has been no previous work on object detection with PG towers, may limited by the lack of available datasets.

\textbf{Segmentation of overhead imagery.}  The second step of \textit{GridTracer} relies on a deep learning model to automatically segment PG lines in the overhead imagery.  Although a variety of segmentation models have been employed on overhead imagery, we focus on two models: the U-net model, and StackNetMTL. The U-net model employs an encoder-decoder structure with skip connections between the encoder and decocder to maintain fine-grainned imagery details to support accurate pixel-wise classifications \cite{ronneberger2015u, Chen2017a}.  The U-Net \cite{ronneberger2015u} and subsequent variants (e.g., Ternaus models \cite{Iglovikov2018}) have recently yielded top performance on the Inria (2018) \cite{Maggiori2017}, DeepGlobe (2019) \cite{demir2018deepglobe}, and DSTL (2018) \cite{Iglovikov2017} benchmarks for object segmentation in overhead imagery.  In addition to their benchmark success, PG lines are very small (sometimes just a single pixel wide) and therefore we hypothesize that fine-grained features will be crucial to detect PG lines.  

The StackNetMTL \cite{batra2019improved} is a recent model that achieved state-of-the-art performance specifically for the segmentation of roads in overhead imagery.   Like PG infrastructure, roads exhibit weak local visual features (though to a lesser extent than transmission lines), and considering the visual features over a larger context can improve recognition \cite{batra2019improved}.  The StackNetMTL, for example, also trains a model to predict both the label of a pixel (i.e., road, or not) and its likely road orientation.  The joint learning process helps the network to better learn the connectivity information, and can also be applied to PG lines in a similar fashion.  For these reasons, and its state-of-the-art performance, we explore the  StackNetMTL for PG line segmentation.   

\textbf{Graph extraction from overhead imagery.} Once the PG towers are identified, we can infer a graphical representation of the grid (i.e., a map), with PG towers as nodes, and PG lines as edges; this is the final step of \textit{GridTracer}.  To our knowledge, the most closely related problem to ours is road mapping (e.g., \cite{batra2019improved, Chaudhuri2015, Mnih2010}. Historically, road mapping was treated largely as a segmentation problem \cite{Mnih2010, Marcu2016, Qian2017}, however, in recent years road mapping has also been formalized as a geospatial graph inference problem, in which road intersections are graph nodes and any intervening roadway are treated as graph edges. Two recent and well-cited models for road graph extraction have recently been proposed: \textit{RoadTracer} \cite{Bastani2018} and \textit{DeepRoadMapper} \cite{mattyus2017deeproadmapper}.  Unfortunately, neither of these methods is directly applicable to the PG mapping problem.  The most immediate challenge is the way in which these two methods create graph nodes: \textit{RoadTracer} places nodes at regular spatial intervals without any regard for any local visual cues, and \textit{DeepRoadMapper} assigns graph nodes at any location where a road segment (identified by a segmentation model) has a discontinuity.  This leaves both methods without any clear mechanism to enforce graph nodes to reside on PG towers, and therefore these methods are not directly applicable to PG mapping (e.g., as baselines for comparison).  For these reasons, we develop \textit{GridTracer} as the first approach that is applicable to the PG mapping problem, and we propose it as a problem-specific baseline approach, upon which future methods can be developed and compared.  


\section{The Power Grid Imagery Dataset}
\label{sec:pg_dataset}
The PG imagery dataset consists of 264 $km^{2}$ of overhead imagery collected over three distinct geographic regions: Arizona, USA (\textit{AZ}); Kansas, USA (\textit{KS}) and New Zealand (\textit{AZ}).   Some basic statistics of the dataset are presented in Table \ref{table:stats}, where the PG infrastructure statistics are derived from human annotations.  We chose these diverse geographic regions so that we (and future users) could demonstrate the validity of any PG mapping approaches across differing geographic settings. 

Although our dataset includes both 0.15m and 0.3m resolution imagery, we resampled all of the imagery to 0.3m for our experiments.  This was done, in part, to maintain consistency of testing results.  A second reason was to enhance the practical relevance of our results; while it is likely that utilizing higher resolution imagery would yield greater PG mapping accuracy, 0.15m resolution imagery is only available via aerial photography, whereas 0.3m imagery is available from satellites (e.g., Worldview 2 and 3 satellites).  Satellite-based imagery offers much greater geographic coverage and imaging frequency compared to aerial photography, while also being less expensive. Our aim here is to explore this problem for imagery that could ultimately support applications across the globe, including areas currently transitioning to electricity access.  By employing 0.3m resolution imagery we will better support these objectives.  


\begin{table}
\begin{center}
    \caption{Summary of power grid annotation imagery}
    \label{table:stats}
    \begin{tabular}{l||l|l|l|l}
    \hline\noalign{\smallskip}
    Region & Area ($km^2$) & \# towers & \# other towers & \# lines\\
    \noalign{\smallskip}
    \hline
    Arizona, USA & 108 & 595 & 23 & 503 \\
    Kansas, USA & 83 & 813 & 102 & 712 \\
    New Zealand & 73 & 1998 & 269 & 1729 \\
    \hline
\end{tabular}
\end{center}
\end{table}

\subsection{Ground truth representation}
\label{sec:gt_representaton}
There are two major classes of objects that we annotated in the imagery: towers and lines.  For the purpose of PG mapping, we need to precisely localize each PG tower in the imagery, as well as provide information about its shape and size to support the training of object detection models.   Therefore each tower was annotated with a bounding box (i.e., a recangle), which is parameterized by a vector $t =(r,c,h,w)$ where $(r,c)$ encodes the pixel \textit{location} of the top left corner of the box (the row and columns of the corresponding pixel), and $(h,w)$ encodes the height and width (again, in pixels) of the rectangle.  

Let $T$ denote the set of all $t$-vectors in the ground truth of the dataset, which specifies the location of each tower. Given $T$, the PG lines can be represented very succinctly by observing that PG lines always form straight line segments between the centroids of PG tower bounding boxes.  Therefore, the precise visual extent of a particular PG line can be accurately inferred simply by knowing which two towers are connected by that line.   Therefore, we can succinctly represent the PG lines in the imagery by an adjacency matrix, $A$, where $A_{ij}=1$ indicates that there is a connection (a power line) between the $i^{th}$ and $j^{th}$ towers in $T$, and $A_{ij}=0$ otherwise.  Adjacency matrices are commonly used to succinctly represent graphs, and therefore the PG is naturally conceptualized as a graph.  However, the nodes in the PG graph are each associated with a geospatial location, distinguishing them from generic graphs in mathematics.  Therefore, we refer to the PG as a geospatial graph, which is characterized by a set of node locations as well as an adjacency matrix. 

\subsection{Annotation details}
All ground truth labels were acquired via manual human annotation of the color overhead imagery. The imagery was split into non-overlapping sub-images, termed "tiles", approximately $5k\times 5k$ in size.  Each tile was manually inspected and annotated using a software tool especially designed for the rapid annotation of overhead imagery.  The tool allows users to choose between two \textit{primary} types of annotation: rectangles for towers (label “T”) and line segments for lines (“L”).  These annotations are then used to generate the $T$ and $A$ ground truth matrices discussed in Section \ref{sec:gt_representaton}.  One separate $T$ and $A$ matrix was generated for each image tile, so that each tile could be processed and scored independently.   

\begin{figure}[h]
\centering
\includegraphics[width=0.5\textwidth]{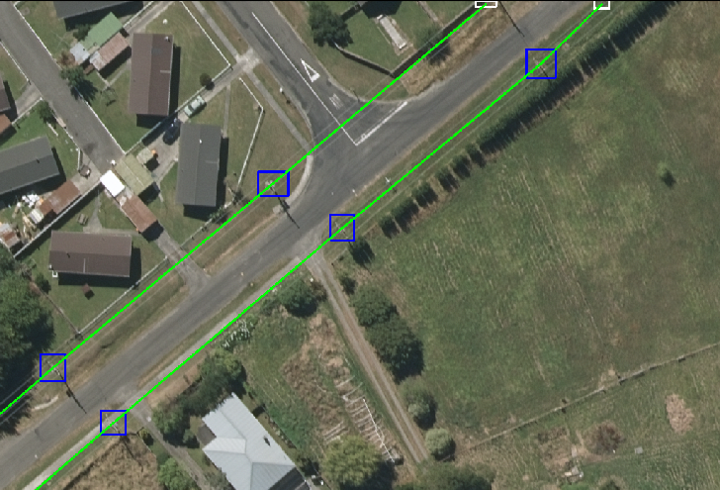}
\caption{An example of annotation. The green lines represent the power line and blue box represent the transmission tower. The white box represents the edge node, which is not a physically real object. It is annotated for the completeness of power line connectivity across tiles.}
\label{fig:annotate_en}
\end{figure}

Rectangular annotations were drawn so that they enclosed the entire physical extent of each tower, excluding shadows. Examples of tower annotations are provided in Fig. \ref{fig:annotate_en} as blue squares. In a small subset of cases the annotators were uncertain about whether a tower is an electricity tower or some other type of tower e.g., streetlights. The annotators were instructed in those cases to assign an “Other Tower” category (“OT”).  “OT” annotations are still included in the ground truth matrices as graph nodes however, the "OT" indicator is included in the ground truth meta data so that users can decide how to use these towers.  In this work we use "OT" towers for training since many of these objects look similar to PG towers and the models may benefit from the additional training imagery.  However, we exclude "OT" labels from evaluation because we want to measure performance only on true PG infrastructure.   

Annotators were instructed to draw line segments between any two towers that were connected by a power line. In such cases, a line segment was drawn from the center of the first tower’s rectangle to the center of the second tower's rectangle. Examples of line segment annotations are provided by the green lines in Fig. \ref{fig:annotate_en}.  

It sometimes occurs that PG lines connect towers in two neighboring tiles. This potentially creates substantial additional complexity when annotating and processing the imagery.  This can also substantially slow down the processing of imagery with deep learning models because large image tiles may not fit into the memory of graphics processing units.  In order to circumvent these potential problems, we created artificial "Edge Nodes" (EN) that were placed at any location where a PG line crossed the boundary of an image tile.  An example EN node is shown in Fig. \ref{fig:annotate_en}.  This formalism allows us to maintain a fully self-contained graph representation of each image tile so that each tile can be processed separately.  Similar to the "OT" nodes, the EN nodes were included in all ground truth as graph nodes.  However, because EN nodes do not actually represent PG towers, we do not use "EN" nodes for training or evaluating tower detection or PG graph inference output.   

Each annotator was trained to recognize PG towers (including the OT and EN designations) and lines using two especially challenging tiles of imagery that were identified by our team.  To ensure overall quality and consistency in the dataset, each annotator's training annotations were reviewed for accuracy before that annotator was permitted to annotate more tiles.  We also note that annotators were asked to annotate any substations ("SS") that were present in the imagery. These substations were not included in our experiments, however for completeness, we include their annotation details in the Appendix.    

\subsection{Dataset characteristics and analysis}
\label{sec:ds_stats}
 In this section, we present qualitative and quantitative analyses of the dataset to (i) illustrate the diversity of the dataset, as well as (ii) provide useful information for algorithm development and analysis of our experimental results.  Basic statistics regarding the dataset are presented in Table \ref{table:stats}. From these basic statistics we can see that there are substantial differences between the three regions.  For example, New Zealand has the highest density (per unit area) of PG infrastructure – by a large margin - while Kansas has a greater density than Arizona.  New Zealand also has the greatest number of line connections per tower.  

\begin{figure}
    \centering
    \includegraphics[width=0.5\textwidth]{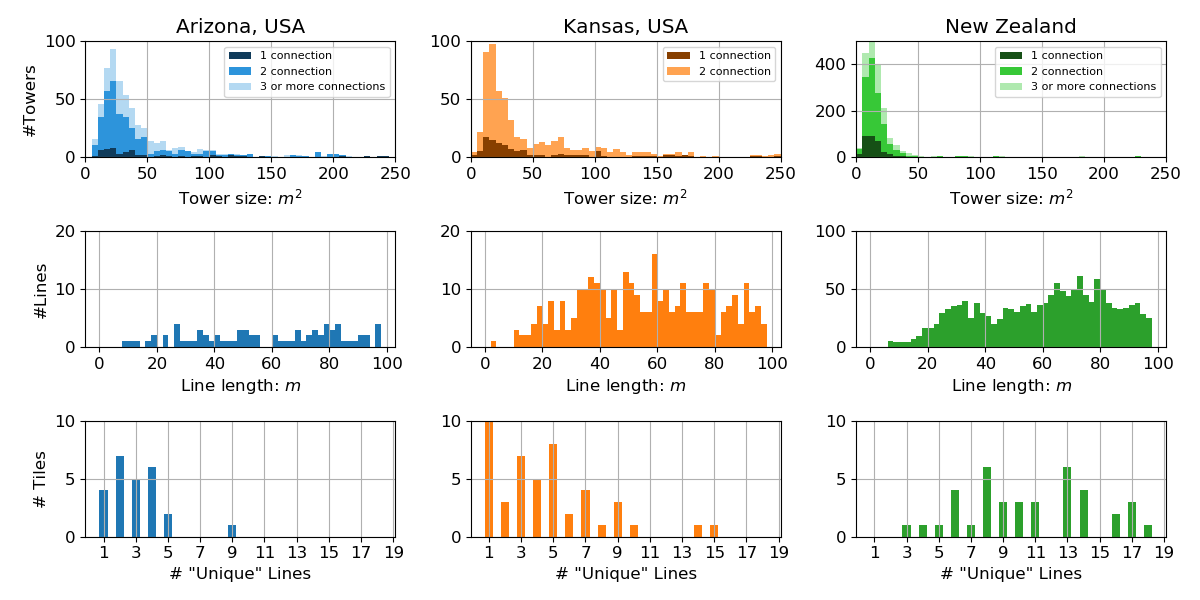}
    \caption{\textit{Top Row}: Stacked histogram of the tower (‘T’) sizes. Dark color bars mean towers only connected to two other towers. Light color bars mean towers connected to three or more towers. \textit{Middle Row}: Distribution of the lengths power line length between two connected towers. \textit{Bottom Row}: Number of unique power line angles categorized in 20 groups. More unique angles mean more complex connection pattern. Note the scale of the y axis is not the same for New Zealand in the first and third rows.}
    \label{fig:dststs}
\end{figure}


Fig. \ref{fig:dststs} presents several other useful statistical features associated with the dataset, stratified by location.  We briefly summarize some of the main observations. The first row indicates that New Zealand has substantially smaller towers than the US locations on average. All three plots in the top row also show that the towers in all three regions are usually connected to two towers, with a small number of towers having three or more connections, especially in Arizona. The second row indicates that there is a wide, but relatively similar, distribution of line lengths across all three locations. This is a useful distribution for limiting the potential towers connected to a given tower (e.g., we need not consider towers further than 100 meters away). Furthermore, we observe New Zealand has substantially more power lines compared to the other three regions. Finally, in row three, we create a crude measure for the complexity of the PG (per unit area); we count the number of unique line angles within each tile, using 18 discrete possible angle bins. A histogram of the number of unique line angles, per tile, is created for each location.  The results indicate that the PG in New Zealand is substantially more complex, on average, than the others, since the histogram suggests that power lines in New Zealand have various different orientations. This analysis provides important information such as the two regions in US are relatively similar, except Kansas has more power line connections and the grid pattern is slightly more complex. The New Zealand region has substantially more complex PG compared to the other two regions. 


\subsection{Annotation Quality Assessment}

To assess the quality of our annotations, we chose approximately 10\% of the image tiles randomly (distributed equally across the three regions) and produced two independent sets of annotations for each tile: one set made by each of two unique annotators.  We then computed the agreement between the annotations made by the two annotators.  Evaluating annotator agreement is a common strategy for assessing the quality of annotations of machine learning training data \cite{cordts2016cityscapes, lin2014microsoft}.  Two towers were declared as matches if their centroids were within $3m$ of one another; two lines were declared as matches if both ends of each line segment matched with both ends of another line segment, within $3m$ in each case.

\begin{figure}
    \centering
    \includegraphics[width=0.48\textwidth]{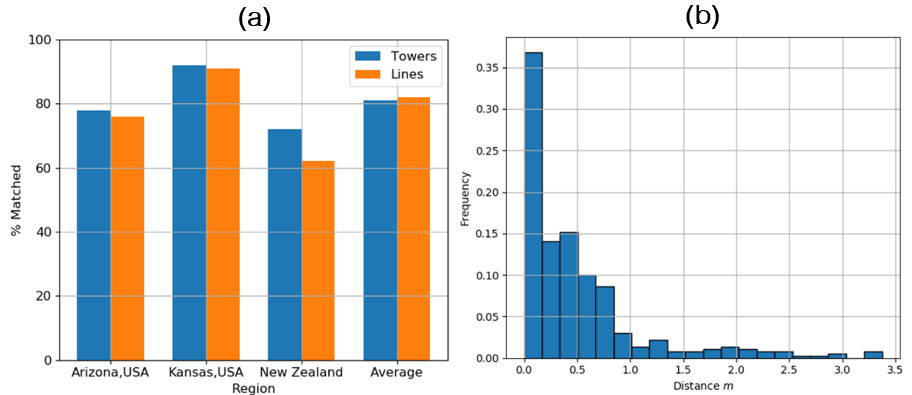}
    \caption{Quantitative measure of agreement between annotators. (a) Percentage of tower centroids being annotated within $3m$ range of different annotators. (b) Histogram of centroid distance between different annotators.}
    \label{fig:annotator_agreement}
\end{figure}

Fig. \ref{fig:annotator_agreement}(a) summarizes the agreement between the annotators.  Tower annotations exhibit a 70-90\% agreement across the three locations. Line annotations exhibit slightly lower agreement for each location because line agreement depends upon first correctly-identifying the tower locations.  Overall, the results show strong agreement among the annotations, suggesting that consistent PG mapping results may be feasible, given a sufficiently-sophisticated recognition model (e.g., approximated by a human analyst in this case).  Similarly, these results also suggest our annotations are suitable for measuring the recognition accuracy of automatic models.  For example, we expect that a sufficiently-sophisticated automatic recognition model should be capable of achieving a 70-90\% agreement with our annotations.  By contrast, if our human annotations had instead demonstrated very little agreement between annotators (e.g., an approximation of a sophisticated recognition model), then it will be difficult to distinguish between poor detectors and good detectors, suggesting automatic PG mapping may be infeasible.    

\section{\textit{GridTracer}: A Baseline Model for Power Grid Mapping in Overhead Imagery}
\begin{figure*}[ht]
    \centering
    \includegraphics[width=18cm]{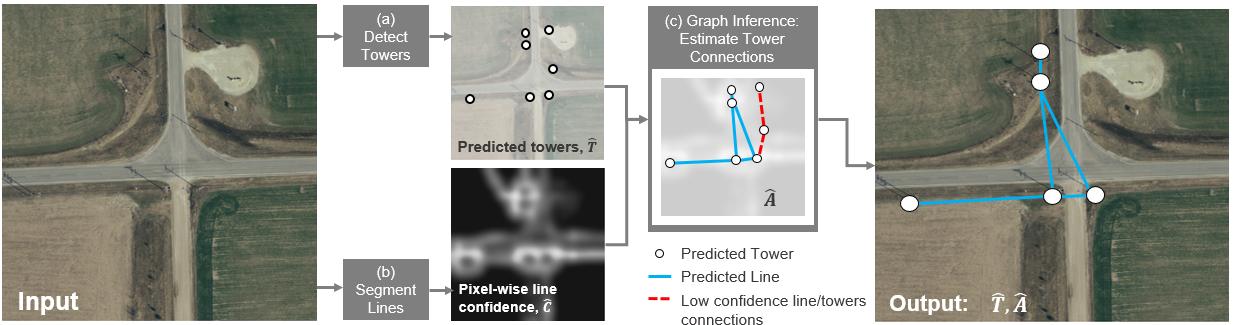}
    \caption{An illustration of our baseline grid mapping algorithm: \textit{GridTracer}. (a) We use an object detection model to infer the locations of PG towers, $\hat{T}$. (b) We use a segmentation model to infer an image, $\hat{C}$, where each pixel intensity indicates the probability that a line exists at that location. (c) Using $\hat{T}$ and $\hat{C}$ we infer the connections between each tower, which is given by an adjacency matrix, $\hat{A}$. The final output of the model is a geospatial graph of the PG characterized by $\hat{A}$ and its associated PG tower locations, $\hat{T}$.}
    \label{fig:system}
\end{figure*}

In this section we present our baseline model \textit{GridTracer} for the PG inference problem. We break down the PG graph inference problem into three sub-problems: tower detection, line segmentation and graph inference. The processing pipeline of \textit{GridTracer} is illustrated in Fig. \ref{fig:system}.

\subsection{Tower Detection}

The goal of tower detection is to predict the centroid of each PG tower. However, our ground truth annotations provide richer information about each tower – full rectangles. This makes it possible to naturally apply and train state-of-the-art object detection models for tower detection.  Due to the challenging nature of tower detection, we focus on maximum accuracy and employ a two-stage (as opposed to one-stage) object detector: the Faster RCNN \cite{ren2015faster}. We trained faster RCNN with Inception V2 \cite{Szegedy2016} on our proposed PG dataset to detect towers. In section \ref{sec:model_ablation} we provide results from an ablation study using different backbone choices and find that the Inception V2 generally yields the best results.  Due to the relatively small size of PG towers in imagery compared to other objects to which object detectors have been applied \cite{lin2014microsoft, everingham2010pascal}, we had to significantly reduce the size of the bounding box anchors to achieve good results.

At inference time, we first extract the raw imagery into $500\times 500$ sub-images. We apply the tower detector on those sub-images and only keep the boxes with a confidence higher than 0.5. Since we are only interested in the location instead of the size of the towers, the centers of the bounding boxes were retained as predictions. We use Non-Maximum Suppression (NMS) \cite{Bodla_2017_ICCV} to remove redundant predictions for nearby bounding boxes. The result of this process is a list of estimated PG tower centroid locations, termed $\hat{T}$, illustrated in Fig. \ref{fig:system}.

\subsection{Line segmentation}
As described in section \ref{sec:introduction}, the PG lines are conceptualized as edges in the PG graph.  As a result, the precise location (e.g., pixel-wise segmentation) of the PG lines are unnecessary for the final output of \textit{GridTracer}, however, having such information is useful to determine whether a line exists.  As a result, as an intermediate step, \textit{GridTracer} employs a state-of-the-art segmentation model to infer all locations throughout the imagery that may indicate PG lines.  The output of this model will then be utilized in the next stage of processing (graph inference) to infer which towers are most likely to be connected by PG lines.    

To extract a segmentation map of PG lines, we employ the StackNetMTL, which has recently achieved success for road segmentation \cite{batra2019improved}.  As discussed in \ref{sec:related}, the StackNetMTL includes a greater visual context when inferring target labels, which we hypothesize may also benefit PG line segmentation.  We provide ablation studies in section \ref{sec:model_ablation} indicating that this is indeed the case.

In order to train our segmentation models, we must create ground truth imagery that indicates which pixels are PG lines (pixel value of one), and which are not (pixel value of zero).  To do this, we use our manual annotations to draw straight lines between the centroids of each pair of connected towers.  Each line is 30-pixels-wide and all pixels in the line are set to a value of one. This width is chosen to ensure that the ground truth labels encompass the real power lines, whose exact locations in the imagery are unknown.  Once trained, we apply StackNetMTL to produce a map of pixel-wise PG line probabilities, termed $\hat{C}$, that is illustrated in Fig. \ref{fig:system}.

\subsection{Graph inference}
The goal of this step is to infer an adjacency matrix, $\hat{A}$,  where $\hat{A}_{ij}=1$ indicates that there is a connection between the $i^{th}$ and $j^{th}$ towers in $\hat{T}$, and $\hat{A}_{ij}=0$ otherwise.  To infer these connections, we will rely upon the output of the PG line segmentation model, $\hat{C}$. Each pixel in $\hat{C}$ indicates the relative likelihood that a power line exists at pixel k.  \textit{GridTracer} will label $\hat{A}_{ij} =1$ if and only if two conditions are met: (i) the distance between tower $i$ and $j$ is less than some user-defined threshold, $d$; and (ii) $S_{ij}\ge\gamma$, where $\gamma$ is a user-defined threshold. $S_{ij}$ is the estimated likelihood that a connection exists between tower $i$ and $j$, based upon integrating the pixel-wise output of a power line segmentation model (StackNetMTL), along the path between the two towers, given by
\begin{equation}
S_{ij}=\frac{1}{|P_{ij}|}\sum{C_{ij}}.
\end{equation}
Here $P_{ij}$ is the set of pixels in the path between the towers, which is a straight-line segment, of width $w$, between the two towers. This simple operation allows us to integrate visual cues well beyond the field-of-view of the segmentation model. Once $S_{ij}$ is obtained, we retain the connection between tower $i$ and $j$ if $S_{ij}$ exceeds a threshold value, $\gamma$, and the distance between the two towers is smaller than a predefined parameter, $d$. In practice, we only consider connections between towers that are within $d$ pixels of one another, dramatically limiting the number of candidate connections we need to consider.

As illustrated in Fig. \ref{fig:system}, the final output of \textit{GridTracer} is a geospatial graph of the PG characterized by $\hat{A}$ and its associated PG tower locations, $\hat{T}$.

\section{Experimental design}
\label{sec:experimental_design}
In this section we describe the major experimental design details.  A major goal of this paper is to establish a benchmark for PG mapping, and therefore we prescribe a proposed data handling scheme for training and evaluating PG mapping models.  We also propose a set of scoring metrics for evaluating models. Finally, we describe the implementation details of \textit{GridTracer}. 
\begin{figure}
    \centering
    \includegraphics[width=0.48\textwidth]{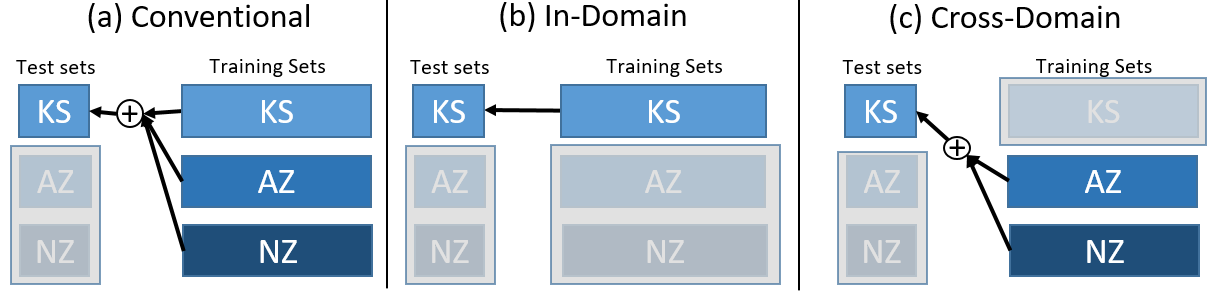}
    \caption{Three different data handling schemes. \textit{(a)}: Train a separate model for each region and evaluate on the corresponding region. \textit{(b)}: Train one model on all regions and test on all regions. \textit{(c)}: Train model on all but one region and evaluate on the held-out region.}
    \label{fig:setup}
\end{figure}

\subsection{Data handling for model training}
\label{sec:data_handling}
We explore three data handling schemes, as illustrated in Fig. \ref{fig:setup}. In all data handling schemes, we use the same subset of the imagery from each location for testing (first 20\% of imagery of each city) so that, regardless of the data handling scheme, the exact same testing dataset is always employed.   Fig. \ref{fig:setup}(a) is the "conventional" data-handling scheme, in which training imagery is available from all testing locations, and models are trained on all available training imagery.  This approach is labeled as the "conventional" because it is commonly-employed on overhead imagery recognition benchmark problems (e.g.,DeepGlobe \cite{demir2018deepglobe}, DSTL \cite{Iglovikov2017}).  For this reason we prescribe this as the primary data handling scheme for our PG mapping benchmark dataset.   Our main results in section \ref{sec:main_results} are obtained using this scheme.  Fig. \ref{fig:setup}(b,c) presents two additional data handling schemes that we utilize in Section \ref{sec:additional_analysis} for further analysis.  We describe the motivation for these designs in Section \ref{sec:additional_analysis}.


\subsection{Scoring: tower detection}
\label{sec:tower_detection_scoring}
As discussed in the introduction, we split the PG mapping problem into two sub-problems: tower detection and tower connection (i.e. line interconnection). 

For tower detection, we adopt the mean average precision (mAP) metric because it is widely-used for object detection tasks (e.g.,  \cite{everingham2010pascal,lin2014microsoft,lin2014microsoft}).  mAP is computed by first assigning a label to each predicted box, $\hat{b} \in \hat{B}$, indicating whether it is a correct detection, or a false detection.  This label is based upon whether a given predicted box achieves a sufficiently high IoU with at least one ground truth box, $b \in B$. Mathematically, we have  
\begin{equation}
    \label{eq:iou_linking}
    l_{i} = 1[ \max_{b_{j} \in B} \; IoU(\hat{b}_{i},b_{j}) > \tau]
\end{equation}
where $l_{i}$ is the label assigned to the $i^{th}$ predicted bounding box, and $\tau$ is a user-defined threshold.  In this work we utilize an alternative matching criteria that depends instead on the the distance between the centroid of the predicted and ground-truth boxes. Mathematically, we have
\begin{equation}
\label{eq:distance_linking}
    l_{i} = 1[ \max_{b_{j} \in B} \; d(\hat{b}_{i},b_{j}) > \tau]
\end{equation}
where $d$ is the distance between the centroids of the bounding boxes. We term this modified metric distance-based mAP ($DmAP$).  We also utilize Eq. \ref{eq:distance_linking} for our PG graph scoring metric (discussed next in Section \ref{sec:grid_scoring_metrics}), since it also requires linking predicted and ground truth towers. 

\begin{figure}[h]
\centering
\includegraphics[width=0.4\textwidth]{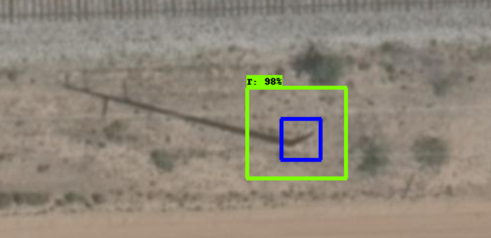}
\caption{A tower detector prediction example. Where the blue box is the annotated ground truth and green box is the prediction with confidence score on the top left.}
\label{fig:map_demo}
\end{figure}

We rely upon eq. \ref{eq:distance_linking} for linking because, in PG mapping, we are primarily concerned with the accuracy of the \textit{locations} (e.g., centroids) of the predicted towers rather than their precise shape and size.  Furthermore, we find that our $mAP$ scores for PG tower detection are often very low, even while our $DmAP$ scores are high, indicating that IoU may indicate a poor prediction even while the location of the predicted tower is accurate.  We present results in Section \ref{sec:additional_analysis} indicating that this is indeed the case for our benchmark dataset and our models.  Fig. \ref{fig:map_demo} presents a typical example of such a scenario.   

As a result of this problem we utilize the linking criterion in eq. \ref{eq:distance_linking} for all of our benchmark scoring metrics, unless otherwise noted.  When using eq. \ref{eq:distance_linking}, we use $\tau=3m$ because it reflects the variability of human annotations made over the same towers. As illustrated in Fig. \ref{fig:annotator_agreement} (b), centroids of human annotations fall within $3m$ of each other roughly 99\% of the time.   

\subsection{Scoring: power grid inference}
\label{sec:grid_scoring_metrics}
For the evaluation of tower connections, our goal is to reward true power line connections between real towers, and penalize any predictions that are incorrect  (including those between falsely detected towers). There is no previously defined metric for assessing the accuracy of PG network predictions, and therefore we propose one here to reflect our goals. One existing metric that captures these goals well is the SGEN+ proposed in \cite{Yang2018} to score predictions of graphical structures in scene understanding problems. This metric includes a “recall” measure, indicating the proportion of true graph connections (i.e., power lines) identified by a model, but no "precision" metric. This is because in \cite{Yang2018}, the authors are trying to study the relationship between the object pairs contained in the image and the annotators are not able to use language to describe all relationship between object pairs. Therefore, the "precision" metric is not included since the authors do not want to penalize models for predicting relationship that is not described by the annotators. However, in the PG mapping task, the connection relationship is well defined by the power lines, therefore we propose the scoring metric as:
$$
R=\frac{C(T)+C(L)}{N_{truth}}, P=\frac{C(T)+C(L)}{N_{pred}},F1=2\times\frac{R\times P}{R+P}
$$ 
Here, $R$ and $P$ represents recall and precision, respectively. $C(\cdot)$ is the counting operation and $T$, $L$ stands for correctly recognized nodes (towers) and edges (lines), respectively. $N_{pred}$ represents the total number given by predictions and $N_{truth}$ represents the total number of objects and relationships given by the ground truth.    As discussed in Section \ref{sec:tower_detection_scoring}, we use the criterion in equation \ref{eq:distance_linking} to determine when detected towers match ground truth towers. 

\subsection{Implementation details of \textit{GridTracer}}
There are two deep learning models in \textit{GridTracer}: a tower detector and a tower connector (for identifying power line connections). We train each of these two components separately. We train the faster RCNN tower detectors (one for each set of training data) and evaluate them on $500\times 500$ uniformly extracted sub-images from our raw input imagery. We use anchors with area of \{$10^2$,$25^2$,$50^2$,$100^2$,$200^2$\} pixels, and aspect ratio of \{0.5,1.0,2.0\}. These boxes are much smaller than the original RCNN \cite{Yang2018}, inspired by similar work with overhead imagery in \cite{pang2019mathcal}. The first two anchor sizes were chosen specifically to capture the small-scale towers especially in New Zealand. During training, we augmented the training data using both random horizontal and vertical flips as well as 90, 180, and 270 degree rotations. For all of the experiments, we train the models with a batch size of 5 for 50,000 iterations using the aforementioned training image partition method. We adapted a manual learning rate scheduler which uses a learning rate of $3e^{-3}$ for the first 10k steps and drop by a factor of 0.1 after every 10k steps.

At inference time, after the locations of the towers are predicted, we use \textit{gridTracer} with the predicted tower locations to predict the PG. For the hyper-parameters, we set $\gamma=0.2, d=600m, w=9$. We present ablation results for selecting these parameters in Section \ref{sec:grid_ablation}.

\subsection{Human-level performance estimation}
In order to aid our analysis of the PG mapping problem, and GridTracer, we estimate the level of performance that a human annotator may achieve on our dataset.  Human-level performance is often used as a benchmark for visual recognition tasks \cite{dodge2017study, Netzer2011, stallkamp2012man, choi2013human} because humans often (though not always) achieve strong performance on visual recognition techniques, and furthermore, a sufficiently-sophisticated automatic approach should be able to achieve the same performance as a human.  Therefore if an automatic approach does not reach human-level performance, it implies that the recognition model may be making incorrect or incomplete assumptions, and further investment might yield greater performance.  Similarly, if human performance is poor on a given task, it may indicate that a visual task is difficult or even infeasible. We will use human-level performance to ensure the overall feasibility of our PG mapping problem, and assess the relative performance of \textit{GridTracer}.   

As discussed in Section \ref{sec:experimental_design}, to estimate human-level performance we randomly sampled 20\% of the imagery from each of our three geographic regions, and had them annotated by a second group of human annotators.  Then we treated these annotations as predictions, and assessed their accuracy using the same tower detection and graph inference metrics that we apply to \textit{GridTracer}.  In Section \ref{sec:main_results} we will compare \textit{GridTracer} to our human annotators on the same 20\% subset of imagery.

\section{PG mapping benchmark results}
\label{sec:main_results}
In this section, we present the performance of \textit{GridTracer} using the "conventional" data handling scheme illustrated in Fig. \ref{fig:setup}(a).  As discussed in Section \ref{sec:experimental_design}, this scheme has been employed in numerous recent benchmark for recognition in overhead imagery (e.g.,DeepGlobe \cite{demir2018deepglobe}, DSTL \cite{Iglovikov2017}).  Although \textit{GridTracer} is composed of three steps, here we present the two metrics of our PG mapping benchmark: (i) PG tower detection, and (ii) PG graph inference. We provide line segmentation results, along with other analyses, in Section \ref{sec:additional_analysis}. 

The results here with \textit{GridTracer} represent the first results using an automatic recognition algorithm for both transmission and distribution mapping in overhead imagery, and therefore they represent a baseline upon which other approaches can build. However, because this is a new problem, it is difficult to evaluate (i) the relative success of \textit{GridTracer} and (ii) how much better we could expect to perform with further research?  To address these important questions we estimate human-level performance for this problem, and compare it with \textit{GridTracer} on the same 20\% subset of our testing dataset. Please see Section \ref{sec:experimental_design} for methodological details.  The results of \textit{GridTracer} and the human-level performance will be presented on the 20\% subset of our testing dataset below, in addition to \textit{GridTracer}'s performance on the full testing dataset.

\subsection{Tower detection}
\label{sec:tower_detection_performance}

The PG tower detection results for \textit{GridTracer} are presented in Table \ref{table:tower_detection_results}.  The results indicate that the DmAP score is roughly 0.61 on average across our three testing regions, indicating that almost one out of every two detected towers is false, however, the level of performance varies substantially across regions.  In Arizona, for example, only one in four detected towers is false, while it is closer to one in two in the other regions.  We hypothesize that this difference may be caused by the differences in the background. The shadows, which is usually a critical feature for detectors, of the towers in Arizona, as shown in Fig. \ref{fig:visualization_of_gridtracer_results} top, usually blends with the nearby bushes. This makes the shadows more difficult to recognize, and therefore leads to a decrease in the detection performance.

To provide a reference point for judging the results of \textit{GridTracer}, we can review the results in rows two and three, comparing the performance of \textit{GridTracer} and humans over the same 20\% of our testing data.  First, we note that the performance of \textit{GridTracer} in row one and three are similar, implying that our 20\% testing subset is relatively representative of the full testing dataset.  With this in mind, human annotators achieve a $DmAP$ of 0.86 on average across the three regions, indicating that nearly nine of ten predicted towers will be correct.  Although the level of performance needed to support energy-related decision-making and research will vary, these results provide both energy and computer vision researchers with an estimate of the accuracy that should be achievable with a baseline recognition model, given sufficient development.    

Furthermore, these results help us understand the degree to which \textit{GridTracer} can be improved, and possibly how it can be improved.  Although \textit{GridTracer} relies upon a state-of-the-art object detection model, human performance is substantially better, and more consistent across each of the geographic regions, suggesting significant improvements can be made. As discussed in Section \ref{sec:introduction}, modern DNNs rely primarily upon local visual cues to detect objects.  The substantial performance advantage of humans suggests that they are therefore likely to be using additional cues to identify towers.  We hypothesize that such cues may include the integration of non-local visual features, or exploitation of the known topology/structure of the PG. For example, it may be possible to infer the presence of a PG tower if its inclusion results in a more probable PG topology, even if there are limited visual cues for the tower itself.   

\begin{table}[h]
\begin{center}
    \caption{Tower detection performance for \textit{GridTracer} and human annotators using the $DmAP$ metric (higher better).  Parenthesis indicates the subset of testing data over which each method was scored.  We train \textit{GridTracer} using the "conventional" data handling scheme in Fig. \ref{fig:setup} (a)}
    \label{table:tower_detection_results}
    \begin{tabular}{l||c|c|c|c}
    \hline \hline
     & Arizona & Kansas & New Zealand & Average \\
    \hline \hline
    \textit{GridTracer} (100\%) & 0.73 & 0.54 & 0.55 & 0.61 \\
    \hline \hline
    Human (20\%)  & 0.92 & 0.88 & 0.79 & 0.86 \\
    \hline
    \textit{GridTracer} (20\% ) & 0.72 & 0.55 & 0.49 & 0.59 \\
    \hline \hline
\end{tabular}
\end{center}
\end{table}


\subsection{PG graph inference}

The PG graph inference results for \textit{GridTracer} are presented in Table \ref{table:grid_results}. We apply a similar analysis here to the one before in Section \ref{sec:tower_detection_performance}.  \textit{GridTracer}'s average F1 score of 0.63 indicates that approximately 63\% of the underlying PG (i.e., towers plus connections) is identified, and that 63\% of the inferred PG infrastructure is correct.  This is roughly similar to \textit{GridTracer's} performance for tower detection alone, although caution must be taken when comparing the results here to those in Table \ref{table:tower_detection_results} due to differences in the scoring metrics.       

Similar to the results for PG tower detection, we find that graph inference scores for \textit{GridTracer} on the 20\% subset are similar to those on the full dataset, indicating that the testing subset is representative of the full dataset.  We also again find that human annotators achieve substantially better performance than \textit{GridTracer}, indicating that further improvements can be made to the PG graph inference approach.  However, it is notable that human performance is lower on the graph inference problem, indicating that (given the $0.3m$ resolution of our imagery), PG graph inference may have a lower achievable performance than tower detection.  As with tower detection, the level of accuracy in these output data that is needed to support energy-related decision-making and research will vary.  These results may therefore provide valuable insights regarding the potential utility of PG mapping in overhead imagery for different applications. 

\begin{figure*}[ht]
    \centering
    \includegraphics[width=18cm]{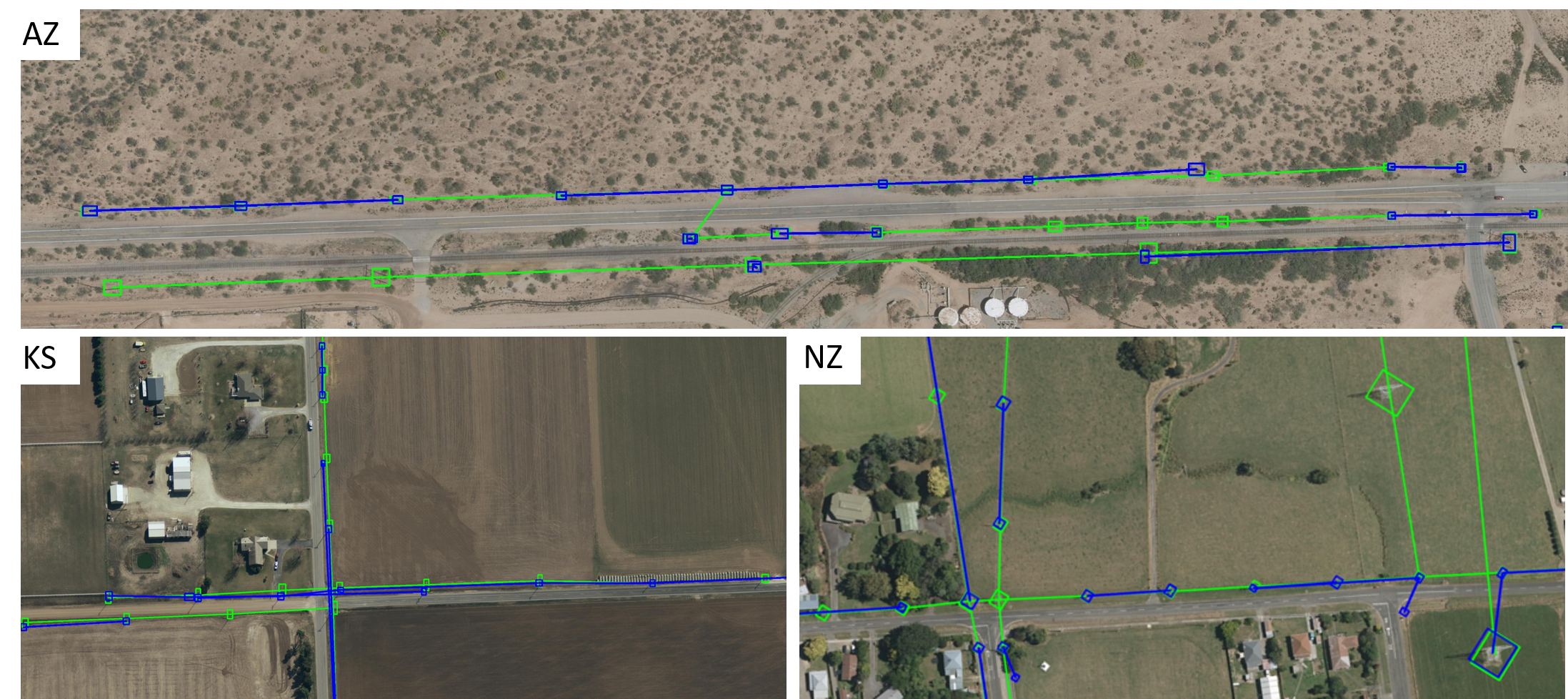}
    \caption{PG inference results of \textit{GridTracer}. Ground truth PG towers and lines are presented in green, while predicted PG towers and lines are presented in blue.}
    \label{fig:visualization_of_gridtracer_results}
\end{figure*}


\begin{table}[h]
\begin{center}
    \caption{PG graph inference performance for \textit{GridTracer} and human annotators using the F1 graph metric (higher better).  Parenthesis indicates the subset of testing data over which each method was scored.  We train \textit{GridTracer} using the "conventional" data handling scheme in Fig. \ref{fig:setup} (a)}
    \label{table:grid_results}
    \begin{tabular}{l||c|c|c|c}
    \hline
     & Arizona & Kansas & New Zealand & Average \\
    \hline \hline
    \textit{GridTracer} (100\% data) & 0.68 & 0.59 & 0.61 & 0.63 \\
    \hline \hline
    Human (20\% data)  & 0.75 & 0.88 & 0.68 & 0.77 \\
    \hline
    \textit{GridTracer} (20\% data) & 0.73 & 0.59 & 0.61 & 0.64 \\
    \hline
\end{tabular}
\end{center}
\end{table}

Fig. \ref{fig:visualization_of_gridtracer_results} presents a visualization of the PG inferred by \textit{GridTracer} compared to the ground truth in each of our three geographic test regions.  These results illustrate the various types of errors that are made by \textit{GridTracer}, such as undetected towers, which also necessarily result in one or more (usually more) undetected PG lines.  Although \textit{GridTracer} finds the majority of towers and connections, the effects of these errors is that the inferred PG graph \textbf{is not} consistent with common PG topology: e.g., towers are connected to two (or more) PG lines, and there are no disjoint subgraphs arising from a single missing connection or tower.  By contrast, human annotations almost always fit these real-world constraints and we hypothesize that humans leverage apriori knowledge about the PG to infer the presence of PG infrastructure even when weak or non-existent visual cues are present.  Given its current design, \textit{GridTracer} does not exploit, or impose upon its predictions, most of these topological cues, and we believe this is an important direction for future work. 

\section{Additional Analysis}
\label{sec:additional_analysis}
In this section we present additional experimental results and analysis that provide further insights into the performance and design of \textit{GridTracer}.

\subsection{Object detection scoring metrics} 

Recall in Section \ref{sec:tower_detection_performance} we argued that the $mAP$-based metric tends to disagree with our $DmAP$ metric, justifying our adoption of the $DmAP$ scoring metric for our PG mapping benchmark.  In this sub-section we provide experimental results supporting this claim.  In Table \ref{table:tower_result} we compare the $DmAP$ metric to two commonly-used $mAP$-based metrics on our benchmark testing dataset: $mAP_{0.5}$ and $mAP_{0.75}$.  The subscript of each $mAP$ score denotes the value of $\tau$ in eq. \ref{eq:iou_linking}.  We see that the $DmAP$ score is higher (indicating better performance) compared to both of the $mAP$-based scores. This suggests (as argued in Section \ref{sec:tower_detection_performance}) that the shape and size of \textit{GridTracer's} predicted bounding boxes often meet the centroid-based $DmAP$ score - the metric of our primary concern in PG mapping - even if they do not meet the IoU-based $mAP$ score.   

\begin{table}[h]
\begin{center}
    \caption{Performance of the tower detection component using the $DmAP$ metric (higher better) of \textit{GridTracer} with the "conventional" data handling scheme in Fig. \ref{fig:setup} (a)}
    \label{table:tower_result}
    \begin{tabular}{l||c|c|c|c}
    \hline 
    Metric & Arizona & Kansas & New Zealand & Average \\
    \hline \hline
    $DmAP$  & 0.73 & 0.54 & 0.55 & 0.63  \\
    \hline
    $mAP_{0.75}$ & 0.09 & 0.02 & 0.08 & 0.06 \\
    $mAP_{0.5}$ & 0.60 & 0.44 & 0.51 & 0.52 \\
    \hline \hline
\end{tabular}
\end{center}
\end{table}

\subsection{Object detector encoder comparisons}
\label{sec:model_ablation}

In this section we report the performance of three pre-trained backbone networks, or "encoders", that we considered for inclusion in \textit{GridTracer}'s tower detection model.  It has been shown in several fields that large pre-trained encoders can offer performance advantages, including in overhead imagery \cite{Zhou2018a,Iglovikov2018}.  Here we consider three widely-used encoders, in order of their size: ResNet50 \cite{he2016deep}, ResNet101 \cite{he2016deep}, and InceptionV2 \cite{Szegedy2016}.  In Table \ref{table:ablation_tower} we compare the performance of faster R-CNN tower detector models, each using a different encoder, on our PG mapping benchmark task for tower detection.  Among them, ResNet50 yielded the worst performance and the other two relatively larger backbones have significantly better results. This result is consistent with other recent findings \cite{Zhou2018a,Iglovikov2018}, and suggests that a large backbone is beneficial for extracting visual features for PG mapping.

\begin{table}[h]
\begin{center}
    \caption{Tower detection performance in $DmAP$ (higher better) of different backbones with the "conventional" data handling scheme in Fig. \ref{fig:setup} (a)}
    \label{table:ablation_tower}
    \begin{tabular}{l||c|c|c|c}
    Backbone & Arizona & Kansas & New Zealand & Average \\
    \hline \hline
    ResNet50 & 0.45 & 0.48 & 0.54 & 0.49  \\
    ResNet101 & \textbf{0.73} & 0.53 & \textbf{0.59} & \textbf{0.62} \\
    Inception V2 & \textbf{0.73} & \textbf{0.54} & \textbf{0.55} & 0.61 \\
    \hline
\end{tabular}
\end{center}
\end{table}

\subsection{PG line segmentation performance and model comparison} 
\label{sec: line_segmentation_results}
In this section we report the performance of two segmentation models that we considered for inclusion in \textit{GridTracer}.  The first is a UNet model with a ResNet50 \cite{he2016deep} encoder that has been pretrained on the ImageNet \cite{JiaDeng2009}.  Models of this form have recently achieved state-of-the-art performance for segmentation of overhead imagery \cite{Zhou2018a, Iglovikov2018}. We also considered the StackNetMTL model (discussed in Section \ref{sec:related}, that recently achieved state-of-the-art performance on road segmentation. Due to the similarities between our task and road mapping, we hypothesized that StackNetMTL may yield better results.  

To assess these two segmentation models, we employed the intersection-over-union (IoU) metric since it is widely used in recent segmentation benchmark problems (e.g., \cite{Iglovikov2017, demir2018deepglobe}). The results of this experiment are presented below in Table \ref{table:ablation_line}.  As we see, StackNetMTL provides substantially and consistently superior performance compared to the UNet.  As a result of this superior performance, we adopted the StackNetMTL in \textit{GridTracer}.  

\begin{table}[h]
\begin{center}
    \caption{Line performance in $IoU$ (higher better) of different backbones with the data handling scheme in Fig. \ref{fig:setup} (a)}
    \label{table:ablation_line}
    \begin{tabular}{l||c|c|c|c}
    \hline
    Model & Arizona & Kansas & New Zealand & Average \\
    \hline \hline
    UNet & 50.02 & 33.85 & 38.38 & 40.75 \\
    StackNetMTL & \textbf{54.43} & \textbf{40.04} & \textbf{46.70} & \textbf{45.57}\\
    \hline
\end{tabular}
\end{center}
\end{table}


\subsection{Robustness to graph inference hyperparameter settings}
\label{sec:grid_ablation}

The graph inference stage in \textit{GridTracer} algorithm has three hyper parameters: $\gamma$, $d$ and $w$. In Table \ref{table:grid_ablation} we show $GridTracer$'s benchmark performance when varying the value of each of these hyperparameters.  We find that $d=600m$ and $w=9$ yields the best performance among the settings we considered, but we also find that \textit{GridTracer} is relatively robust with respect to their settings. We find that performance is somewhat more sensitive to $\gamma$; if it is set too small then we obtain large numbers of false PG line connections, reducing our performance.  However, performance appears to be insensitive once we use larger values, achieving the best performance when $\gamma=0.2$, and dropping only slightly if we set it higher. Overall we find the model is relatively insensitive to these hyperparameter settings.    

\begin{table}[h]
\begin{center}
    \caption{Performance of \textit{GridTracer} using $DmAP$ (higher better) with data handling scheme in Fig. \ref{fig:setup} (a)}
    \label{table:grid_ablation}
    \begin{tabular}{l||c|c|c|c}
    \hline
    $\gamma$ & Arizona & Kansas & New Zealand & Average \\
    \hline \hline
    $0.1$ & 0.62 & 0.51 & 0.54 & 0.56 \\
    $0.2$ & 0.68 & \textbf{0.59} & \textbf{0.61} & \textbf{0.63} \\
    $0.3$ & \textbf{0.69} & 0.56 & 0.59 & 0.61 \\ \hline\hline
    $d$ & Arizona & Kansas & New Zealand & Average \\
    \hline \hline
    $200m$ & 0.64 & 0.55 & 0.59 & 0.59 \\
    $400m$ & 0.65 & 0.58 & \textbf{0.61} & 0.61 \\
    $600m$ & \textbf{0.68} & 0.59 & \textbf{0.61} & \textbf{0.63} \\
    $800m$ & \textbf{0.68} & \textbf{0.60} & 0.60 & \textbf{0.63} \\
    $1000m$ & 0.66 & 0.59 & \textbf{0.61} & 0.62 \\
    \hline \hline
    $w$ & Arizona & Kansas & New Zealand & Average \\
    \hline \hline
    $7$ & \textbf{0.68} & \textbf{0.59} & 0.60 & 0.62 \\
    $9$ & \textbf{0.68} & \textbf{0.59} & \textbf{0.61} & \textbf{0.63} \\
    $11$ & \textbf{0.68} & \textbf{0.59} & 0.59 & 0.62 \\ \hline
\end{tabular}
\end{center}
\end{table}

\subsection{\textit{GridTracer} performance under varying testing scenarios}
\label{sec: geo_diverse}

In this section we consider the performance of \textit{GridTracer} under less conventional testing scenarios.  Our benchmark testing results were based upon the data handling scheme illustrated in Fig. \ref{fig:setup}(a), which is the conventional approach used in most benchmark problems in overhead imagery.  Here we consider the performance of \textit{GridTracer} when tested using the data handling schemes illustrated in Fig. \ref{fig:setup}(b,c).  These experiments are aimed to address two questions: (a) is training on geographically-diverse imagery beneficial? and (b) how well does \textit{GridTracer} generalize to previously-unseen geographic regions?  

\textit{Is geographically-diverse training data beneficial?} As discussed in section \ref{sec:ds_stats}, the three regions in the PG dataset had significantly different visual characteristics, and somewhat unique PG grid topologies.  Given the unique characteristics of these regions, it is unclear whether it is beneficial to train a single model on all regions simultaneously (the "conventional" scheme), as opposed to training a unique model that is tailored to the characteristics of each geographic region.  We address this question by comparing the performance of \textit{GridTracer} when using data handling schemes (a) and (b) in Fig. \ref{fig:setup}.  In contrast to scheme (a), scheme (b) trains a model separately for each geographic region.  

The results of this experiment are presented in Table \ref{table:ablation_split_b}.  In all three stages, the model trained with data handling scheme (a) generally outperforms the one trained with scheme (b).This suggests that sourcing training data from the same visual domain as the testing data tends to under-perform a more geographically (and thereby visually) diverse pool of training data. 

\begin{table}[h]
\begin{center}
    \caption{\textit{GridTracer} performance with the "conventional" data handling scheme in Fig. \ref{fig:setup}(a) and the "in-domain" scheme in the in Fig. \ref{fig:setup}(b), cross-domain training}
    \label{table:ablation_split_b}
    \begin{tabular}{l||c|c|c|c}
    \hline
      & Azrizona       & Kansas         & New Zealand    & Average        \\ \hline\hline
Data handling & \multicolumn{4}{c}{Tower detection (DmaP)}                       \\ \hline
(a)   & \textbf{0.73}  & \textbf{0.54}  & \textbf{0.55}  & \textbf{0.61}  \\
(b)   & 0.68           & 0.53           & 0.51           & 0.57           \\ \hline\hline
      & \multicolumn{4}{c}{Line segmentation (IoU)}                      \\ \hline
(a)   & \textbf{54.43} & \textbf{40.04} & 45.62          & \textbf{46.70} \\
(b)   & 52.97          & 39.42          & \textbf{46.50} & 46.30          \\ \hline\hline
      & \multicolumn{4}{c}{Graph inference (F1)}                         \\ \hline
(a)   & 0.68           & \textbf{0.59}  & \textbf{0.61}  & \textbf{0.63}  \\
(b)   & \textbf{0.70}  & 0.56           & 0.48           & 0.58          \\ \hline
\end{tabular}
\end{center}
\end{table}

\textit{Generalization to unseen geographies.}  The conventional testing scenario in Fig. \ref{fig:setup}(a) is typical in computer vision research implicitly assumes that labeled training data is available in (or near) every geographic location on which we wish to apply our recognition models.  In practice however it is cumbersome and costly to collect training imagery in each deployment location, and re-train the model with that imagery.  In this section we consider how well \textit{GridTracer} performs when evaluated in novel geographic locations - i.e., locations for which no training imagery is available in the training dataset.  We use the data handling scheme in Fig. \ref{fig:setup}(c) to approximate this realistic scenario by limiting the model to be trained on only two regions and then testing on the third (unseen) region. The results are presented in Table \ref{table:ablation_split_c}, compared to the conventional testing scenario.

In scheme (c) both the tower detection and line segmentation yields poor results. This indicates that the model does not generalize well to novel geographic regions.  This finding is consistent with, and corroborates, other recent findings in the literature indicating that deep learning models do not generalize well to new geographic regions \cite{kong2020synthinel, Maggiori2017}. 


\begin{table}[h]
\begin{center}
    \caption{\textit{GridTracer} performance with the "conventional" data handling scheme in Fig. \ref{fig:setup}(a) and the "cross-domain" scheme in Fig. \ref{fig:setup}(c)}
    \label{table:ablation_split_c}
    \begin{tabular}{l||c|c|c|c}
    \hline
      & Azrizona       & Kansas         & New Zealand    & Average        \\ \hline\hline
Data handling  & \multicolumn{4}{c}{Tower detection (DmaP)}                       \\ \hline
(a)   & \textbf{0.73}  & \textbf{0.54}  & \textbf{0.55}  & \textbf{0.61}  \\
(c)   & 0.06           & 0.15           & 0.05           & 0.09           \\ \hline\hline
      & \multicolumn{4}{c}{Line segmentation (IoU)}                      \\ \hline
(a)   & \textbf{54.43} & \textbf{40.04} & \textbf{45.62}          & \textbf{46.70} \\
(c)   & 7.51          & 20.18          & 7.95 &  11.88          \\ \hline\hline
\end{tabular}
\end{center}
\end{table}

\section{Conclusion}
In this work we proposed a novel approach for collecting power grid information automatically by mapping (i.e., detecting and connecting) transmission and distribution towers and lines in overhead imagery using deep learning. We developed and publicly released a dataset of overhead imagery with ground truth information for a variety of power grids. To our knowledge, this is the first dataset of its kind in the public domain and will enable other researchers to build increasingly effective transmission and distribution grid mapping algorithms. 

We also took the first steps towards tackling the PG mapping problem as well. We developed and evaluated baseline algorithms for two problems: tower detection and identifying tower interconnections through power lines.  In particular, we developed \textit{GridTracer} as baseline approach to solve the PG mapping problem.  We also estimate the ability of human annotators to perform PG mapping, providing future researchers with an estimate of the level of PG mapping accuracy that may ultimately be achievable with a fully-automatic mapping algorithm.  In particular, we found that \textit{GridTracer} does not yet reach human-level PG mapping accuracy, suggesting that further improvements can be made to bridge this performance gap.  Ultimately these results provide a strong foundation for the development of automatic PG mapping techniques, which offer a powerful tool to collect valuable information to support energy researchers and decision-makers.  

\appendices

\section*{Acknowledgment}

The authors would like to thank the Duke University Bass Connections and Data+ programs for their support of this work and for each of the team members who contributed to the construction of the dataset including Qiwei Han, Varun Nair, Tamasha Pathirathna, Xiaolan You, Wendell Cathcart, Ben Alexander, Yutao Gong, Xinchun Hu, Lin Zuo. Additional thanks to Wayne Hu for assistance with the dataset and Sang-Jyh Lin, Jose Luis Moscoso, Andy Yang. This work was supported in part by National Science Foundation Grant no. OIA-1937137. Bohao Huang thanks the Duke University Energy Initiative PhD fellowship, funded by the Alfred P. Sloan Foundation for supporting his work.

\ifCLASSOPTIONcaptionsoff
  \newpage
\fi

\bibliographystyle{IEEEtran}
\bibliography{bibtex/bib/IEEEabrv.bib, bibtex/bib/reference.bib}

\begin{thebibliography}{10}
\providecommand{\url}[1]{#1}
\csname url@samestyle\endcsname
\providecommand{\newblock}{\relax}
\providecommand{\bibinfo}[2]{#2}
\providecommand{\BIBentrySTDinterwordspacing}{\spaceskip=0pt\relax}
\providecommand{\BIBentryALTinterwordstretchfactor}{4}
\providecommand{\BIBentryALTinterwordspacing}{\spaceskip=\fontdimen2\font plus
\BIBentryALTinterwordstretchfactor\fontdimen3\font minus
  \fontdimen4\font\relax}
\providecommand{\BIBforeignlanguage}[2]{{%
\expandafter\ifx\csname l@#1\endcsname\relax
\typeout{** WARNING: IEEEtran.bst: No hyphenation pattern has been}%
\typeout{** loaded for the language `#1'. Using the pattern for}%
\typeout{** the default language instead.}%
\else
\language=\csname l@#1\endcsname
\fi
#2}}
\providecommand{\BIBdecl}{\relax}
\BIBdecl

\bibitem{Nations2018}
U.~Nations, ``{The Sustainable Development Goals Report 2018},'' Tech. Rep.,
  2018.

\bibitem{Alstone2015}
P.~Alstone, D.~Gershenson, and D.~M. Kammen, ``{Decentralized energy systems
  for clean electricity access},'' \emph{Nature Climate Change}, vol.~5, no.~4,
  pp. 305--314, 2015.

\bibitem{Szabo2011}
S.~Szab{\'{o}}, K.~B{\'{o}}dis, T.~Huld, and M.~Moner-Girona, ``{Energy
  solutions in rural Africa: Mapping electrification costs of distributed solar
  and diesel generation versus grid extension},'' \emph{Environmental Research
  Letters}, vol.~6, no.~3, 2011.

\bibitem{Mentis2015}
D.~Mentis, M.~Welsch, F.~{Fuso Nerini}, O.~Broad, M.~Howells, M.~Bazilian, and
  H.~Rogner, ``{A GIS-based approach for electrification planning-A case study
  on Nigeria},'' \emph{Energy for Sustainable Development}, vol.~29, pp.
  142--150, 2015.

\bibitem{DevelopmentSeed2018}
{Development Seed}, ``{Mapping the electric grid using ML to augment human
  tracing of HV infrastructure},'' 2018.

\bibitem{arderne2020predictive}
C.~Arderne, C.~Zorn, C.~Nicolas, and E.~Koks, ``Predictive mapping of the
  global power system using open data,'' \emph{Scientific data}, vol.~7, no.~1,
  pp. 1--12, 2020.

\bibitem{Huang2018b}
B.~Huang, K.~Lu, N.~Audebert, A.~Khalel, Y.~Tarabalka, J.~M. Malof, A.~Boulch,
  B.~L. Saux, L.~Collins, K.~Bradbury, S.~Lefevre, and M.~El-Saban,
  ``{Large-scale semantic classification: outcome of the first year of inria
  aerial image labeling benchmark},'' in \emph{International Geoscience and
  Remote Sensing Symposium}, 2018.

\bibitem{demir2018deepglobe}
\BIBentryALTinterwordspacing
I.~Demir, K.~Koperski, D.~Lindenbaum, G.~Pang, J.~Huang, S.~Basu, F.~Hughes,
  D.~Tuia, R.~Raska, and R.~Raskar, ``{Deepglobe 2018: A challenge to parse the
  earth through satellite images},'' in \emph{2018 IEEE/CVF Conference on
  Computer Vision and Pattern Recognition Workshops (CVPRW)}.\hskip 1em plus
  0.5em minus 0.4em\relax IEEE, 2018, pp. 172--17\,209. [Online]. Available:
  \url{http://arxiv.org/abs/1805.06561}
\BIBentrySTDinterwordspacing

\bibitem{Bastani2018}
\BIBentryALTinterwordspacing
F.~Bastani, S.~He, S.~Abbar, M.~Alizadeh, H.~Balakrishnan, S.~Chawla,
  S.~Madden, and D.~Dewitt, ``{RoadTracer: Automatic Extraction of Road
  Networks from Aerial Images},'' \emph{Computer Vision and Pattern Recognition
  (CVPR)}, 2018. [Online]. Available:
  \url{http://nms.lcs.mit.edu/papers/4023.pdf}
\BIBentrySTDinterwordspacing

\bibitem{Malof2016a}
J.~Malof, K.~Bradbury, L.~Collins, and R.~Newell, ``{Automatic detection of
  solar photovoltaic arrays in high resolution aerial imagery},'' \emph{Applied
  Energy}, vol. 183, 2016.

\bibitem{yu2018deepsolar}
J.~Yu, Z.~Wang, A.~Majumdar, and R.~Rajagopal, ``{DeepSolar: A Machine Learning
  Framework to Efficiently Construct a Solar Deployment Database in the United
  States},'' \emph{Joule}, vol.~2, no.~12, pp. 2605--2617, 2018.

\bibitem{Malof2015}
J.~M. Malof, B.~Li, B.~Huang, K.~Bradbury, and A.~Stretslov, ``{Mapping solar
  array location , size , and capacity using deep learning and overhead
  imagery},'' pp. 1--6, 2015.

\bibitem{li2019scale}
Y.~Li, Y.~Chen, N.~Wang, and Z.~Zhang, ``Scale-aware trident networks for
  object detection,'' in \emph{Proceedings of the IEEE international conference
  on computer vision}, 2019, pp. 6054--6063.

\bibitem{luo2016understanding}
W.~Luo, Y.~Li, R.~Urtasun, and R.~Zemel, ``Understanding the effective
  receptive field in deep convolutional neural networks,'' in \emph{Advances in
  neural information processing systems}, 2016, pp. 4898--4906.

\bibitem{mattyus2017deeproadmapper}
G.~M{\'a}ttyus, W.~Luo, and R.~Urtasun, ``Deeproadmapper: Extracting road
  topology from aerial images,'' in \emph{Proceedings of the IEEE International
  Conference on Computer Vision}, 2017, pp. 3438--3446.

\bibitem{Maggiori2017}
E.~Maggiori, Y.~Tarabalka, G.~Charpiat, P.~Alliez, E.~Maggiori, Y.~Tarabalka,
  G.~Charpiat, P.~Alliez, and C.~Semantic, ``{Can Semantic Labeling Methods
  Generalize to Any City ? The Inria Aerial Image Labeling Benchmark},'' pp.
  3226--3229, 2017.

\bibitem{Matikainen2016}
L.~Matikainen, M.~Lehtom{\"{a}}ki, E.~Ahokas, J.~Hyypp{\"{a}}, M.~Karjalainen,
  A.~Jaakkola, A.~Kukko, and T.~Heinonen, ``{Remote sensing methods for power
  line corridor surveys},'' \emph{ISPRS Journal of Photogrammetry and Remote
  Sensing}, vol. 119, pp. 10--31, 2016.

\bibitem{Ahmad2013}
J.~Ahmad, A.~S. Malik, L.~Xia, and N.~Ashikin, ``{Vegetation encroachment
  monitoring for transmission lines right-of-ways: A survey},'' \emph{Electric
  Power Systems Research}, vol.~95, pp. 339--352, 2013.

\bibitem{Kobayashi2014}
Y.~Kobayashi, G.~G. Karady, G.~T. Heydt, L.~Fellow, and R.~G. Olsen, ``{The
  Utilization of Satellite Images to Identify Trees Endangering Transmission
  Lines},'' vol.~24, no. July 2009, pp. 1703--1709, 2014.

\bibitem{lin2017focal}
T.-Y. Lin, P.~Goyal, R.~Girshick, K.~He, and P.~Doll{\'{a}}r, ``{Focal loss for
  dense object detection},'' in \emph{Proceedings of the IEEE international
  conference on computer vision}, 2017, pp. 2980--2988.

\bibitem{ren2015faster}
S.~Ren, K.~He, R.~Girshick, and J.~Sun, ``{Faster r-cnn: Towards real-time
  object detection with region proposal networks},'' in \emph{Advances in
  neural information processing systems}, vol.~39, no.~6, 2015, pp. 91--99.

\bibitem{he2017mask}
\BIBentryALTinterwordspacing
K.~He, G.~Gkioxari, P.~Doll{\'{a}}r, and R.~Girshick, ``{Mask r-cnn},'' in
  \emph{Proceedings of the IEEE international conference on computer vision},
  2017, pp. 2961--2969. [Online]. Available:
  \url{http://arxiv.org/abs/1703.06870}
\BIBentrySTDinterwordspacing

\bibitem{lin2014microsoft}
T.-Y.~Y. Lin, M.~Maire, S.~Belongie, J.~Hays, P.~Perona, D.~Ramanan,
  P.~Doll{\'{a}}r, and C.~L. Zitnick, ``{Microsoft coco: Common objects in
  context},'' in \emph{European conference on computer vision}, vol. 8693 LNCS,
  no. PART 5.\hskip 1em plus 0.5em minus 0.4em\relax Springer, 2014, pp.
  740--755.

\bibitem{Yang2018a}
\BIBentryALTinterwordspacing
X.~Yang, J.~Yang, J.~Yan, Y.~Zhang, T.~Zhang, Z.~Guo, S.~Xian, and K.~Fu,
  ``{SCRDet: Towards More Robust Detection for Small, Cluttered and Rotated
  Objects},'' pp. 8232--8241, 2018. [Online]. Available:
  \url{http://arxiv.org/abs/1811.07126}
\BIBentrySTDinterwordspacing

\bibitem{Li2019a}
C.~Li, C.~Xu, Z.~Cui, D.~Wang, Z.~Jie, T.~Zhang, and J.~Yang, ``{Learning
  Object-Wise Semantic Representation for Detection in Remote Sensing
  Imagery},'' \emph{Cvprw}, pp. 20--27, 2019.

\bibitem{azimi2018towards}
S.~M. Azimi, E.~Vig, R.~Bahmanyar, M.~K{\"o}rner, and P.~Reinartz, ``Towards
  multi-class object detection in unconstrained remote sensing imagery,'' in
  \emph{Asian Conference on Computer Vision}.\hskip 1em plus 0.5em minus
  0.4em\relax Springer, 2018, pp. 150--165.

\bibitem{ronneberger2015u}
\BIBentryALTinterwordspacing
O.~Ronneberger, P.~Fischer, and T.~Brox, ``{U-Net: Convolutional Networks for
  Biomedical Image Segmentation},'' \emph{International Conference on Medical
  image computing and computer-assisted intervention}, pp. 1--8, 2015.
  [Online]. Available: \url{http://arxiv.org/abs/1505.04597}
\BIBentrySTDinterwordspacing

\bibitem{Chen2017a}
\BIBentryALTinterwordspacing
L.-C. Chen, G.~Papandreou, F.~Schroff, and H.~Adam, ``{Rethinking Atrous
  Convolution for Semantic Image Segmentation},'' 2017. [Online]. Available:
  \url{http://arxiv.org/abs/1706.05587}
\BIBentrySTDinterwordspacing

\bibitem{Iglovikov2018}
\BIBentryALTinterwordspacing
V.~I. Iglovikov, S.~Seferbekov, A.~V. Buslaev, and A.~Shvets, ``{TernausNetV2:
  Fully Convolutional Network for Instance Segmentation},'' 2018. [Online].
  Available: \url{http://arxiv.org/abs/1806.00844}
\BIBentrySTDinterwordspacing

\bibitem{Iglovikov2017}
\BIBentryALTinterwordspacing
V.~Iglovikov, S.~Mushinskiy, and V.~Osin, ``{Satellite Imagery Feature
  Detection using Deep Convolutional Neural Network: A Kaggle Competition},''
  2017. [Online]. Available: \url{http://arxiv.org/abs/1706.06169}
\BIBentrySTDinterwordspacing

\bibitem{batra2019improved}
A.~Batra, S.~Singh, G.~Pang, S.~Basu, C.~Jawahar, and M.~Paluri, ``Improved
  road connectivity by joint learning of orientation and segmentation,'' in
  \emph{Proceedings of the IEEE Conference on Computer Vision and Pattern
  Recognition}, 2019, pp. 10\,385--10\,393.

\bibitem{Chaudhuri2015}
D.~Chaudhuri, N.~K. Kushwaha, A.~Samal, and R.~C. Agarwal, ``{Automatic
  Building Detection From High-Resolution Satellite Images Based on Morphology
  and Internal Gray Variance},'' \emph{Selected Topics in Applied Earth
  Observations and Remote Sensing, IEEE Journal of}, vol.~PP, no.~99, pp.
  1--13, 2015.

\bibitem{Mnih2010}
V.~Mnih and G.~E. Hinton, ``{Learning to detect roads in high-resolution aerial
  images},'' \emph{Lecture Notes in Computer Science (including subseries
  Lecture Notes in Artificial Intelligence and Lecture Notes in
  Bioinformatics)}, vol. 6316 LNCS, no. PART 6, pp. 210--223, 2010.

\bibitem{Marcu2016}
\BIBentryALTinterwordspacing
A.~Marcu and M.~Leordeanu, ``{Dual Local-Global Contextual Pathways for
  Recognition in Aerial Imagery},'' 2016. [Online]. Available:
  \url{http://arxiv.org/abs/1605.05462}
\BIBentrySTDinterwordspacing

\bibitem{Qian2017}
S.~Qian, L.~Xiaoping, and L.~Xia, ``{Road Detection from Remote Sensing Images
  by Generative Adversarial Networks},'' vol. 3536, no.~c, pp. 1--10, 2017.

\bibitem{cordts2016cityscapes}
M.~Cordts, M.~Omran, S.~Ramos, T.~Rehfeld, M.~Enzweiler, R.~Benenson,
  U.~Franke, S.~Roth, and B.~Schiele, ``{The cityscapes dataset for semantic
  urban scene understanding},'' in \emph{Proceedings of the IEEE conference on
  computer vision and pattern recognition}, 2016, pp. 3213--3223.

\bibitem{Szegedy2016}
C.~Szegedy, S.~Ioffe, V.~Vanhoucke, and A.~Alemi, ``{Inception-v4,
  Inception-ResNet and the Impact of Residual Connections on Learning},''
  \emph{Arxiv.Org}, 2016.

\bibitem{everingham2010pascal}
M.~Everingham, L.~{Van Gool}, C.~K.~I. Williams, J.~Winn, and A.~Zisserman,
  ``{The pascal visual object classes (voc) challenge},'' \emph{International
  journal of computer vision}, vol.~88, no.~2, pp. 303--338, 2010.

\bibitem{Bodla_2017_ICCV}
N.~Bodla, B.~Singh, R.~Chellappa, and L.~S. Davis, ``Soft-nms -- improving
  object detection with one line of code,'' in \emph{The IEEE International
  Conference on Computer Vision (ICCV)}, Oct 2017.

\bibitem{Yang2018}
J.~Yang, J.~Lu, S.~Lee, D.~Batra, and D.~Parikh, ``{Graph R-CNN for Scene Graph
  Generation},'' \emph{Proceedings of the European Conference on Computer
  Vision (ECCV)}, pp. 670----685, 2018.

\bibitem{pang2019mathcal}
J.~Pang, C.~Li, J.~Shi, Z.~Xu, and H.~Feng, ``R2-cnn: Fast tiny object
  detection in large-scale remote sensing images,'' \emph{IEEE Transactions on
  Geoscience and Remote Sensing}, vol.~57, no.~8, pp. 5512--5524, 2019.

\bibitem{dodge2017study}
S.~Dodge and L.~Karam, ``A study and comparison of human and deep learning
  recognition performance under visual distortions,'' in \emph{2017 26th
  international conference on computer communication and networks
  (ICCCN)}.\hskip 1em plus 0.5em minus 0.4em\relax IEEE, 2017, pp. 1--7.

\bibitem{Netzer2011}
Y.~Netzer and T.~Wang, ``{Reading digits in natural images with unsupervised
  feature learning},'' \emph{Nips}, pp. 1--9, 2011.

\bibitem{stallkamp2012man}
J.~Stallkamp, M.~Schlipsing, J.~Salmen, and C.~Igel, ``Man vs. computer:
  Benchmarking machine learning algorithms for traffic sign recognition,''
  \emph{Neural networks}, vol.~32, pp. 323--332, 2012.

\bibitem{choi2013human}
J.~Choi, H.~Lei, V.~Ekambaram, P.~Kelm, L.~Gottlieb, T.~Sikora, K.~Ramchandran,
  and G.~Friedland, ``Human vs machine: establishing a human baseline for
  multimodal location estimation,'' in \emph{Proceedings of the 21st ACM
  international conference on Multimedia}, 2013, pp. 867--876.

\bibitem{Zhou2018a}
L.~Zhou, C.~Zhang, and M.~Wu, ``{D-linknet: Linknet with pretrained encoder and
  dilated convolution for high resolution satellite imagery road extraction},''
  \emph{IEEE Computer Society Conference on Computer Vision and Pattern
  Recognition Workshops}, vol. 2018-June, pp. 192--196, 2018.

\bibitem{he2016deep}
K.~He, X.~Zhang, S.~Ren, and J.~Sun, ``Deep residual learning for image
  recognition,'' in \emph{Proceedings of the IEEE conference on computer vision
  and pattern recognition}, 2016, pp. 770--778.

\bibitem{JiaDeng2009}
\BIBentryALTinterwordspacing
{Jia Deng}, {Wei Dong}, R.~Socher, {Li-Jia Li}, {Kai Li}, and {Li Fei-Fei},
  ``{ImageNet: A large-scale hierarchical image database},'' \emph{2009 IEEE
  Conference on Computer Vision and Pattern Recognition}, pp. 248--255, 2009.
  [Online]. Available:
  \url{http://ieeexplore.ieee.org/lpdocs/epic03/wrapper.htm?arnumber=5206848}
\BIBentrySTDinterwordspacing

\bibitem{kong2020synthinel}
F.~Kong, B.~Huang, K.~Bradbury, and J.~Malof, ``The synthinel-1 dataset: a
  collection of high resolution synthetic overhead imagery for building
  segmentation,'' in \emph{The IEEE Winter Conference on Applications of
  Computer Vision}, 2020, pp. 1814--1823.

\end{thebibliography}

%

\end{document}